\documentclass[conference]{IEEEtran}
\IEEEoverridecommandlockouts
\usepackage{cite}
\usepackage{amsmath,amssymb,amsfonts}
\usepackage{algpseudocode}
\usepackage{graphicx}
\usepackage{textcomp}
\usepackage{xcolor}
\usepackage{bm}
\usepackage{subfig}
\usepackage{multirow}
\usepackage{booktabs}

\usepackage[a4paper, total={184mm,239mm}]{geometry}
\def\BibTeX{{\rm B\kern-.05em{\sc i\kern-.025em b}\kern-.08em
    T\kern-.1667em\lower.7ex\hbox{E}\kern-.125emX}}
\newenvironment{sequation}{\begin{equation}\small}{\end{equation}}

\makeatletter  
\newif\if@restonecol  
\makeatother

\usepackage[linesnumbered,ruled,vlined]{algorithm2e}

\begin{document}

\title{Parallel Scheduling Self-attention Mechanism: Generalization and Optimization}

\author{\IEEEauthorblockN{Mingfei Yu}
\IEEEauthorblockA{\textit{The University of Tokyo}\\
yu@cad.t.u-tokyo.ac.jp}
\and
\IEEEauthorblockN{Masahiro Fujita}
\IEEEauthorblockA{\textit{The University of Tokyo}\\
fujita@ee.t.u-tokyo.ac.jp}
}
\maketitle

\begin{abstract}
Over the past few years, self-attention is shining in the field of deep learning, especially in the domain of natural language processing(NLP). 
Its impressive effectiveness, along with ubiquitous implementations, have aroused our interest in efficiently scheduling the data-flow of corresponding computations onto architectures with many computing units to realize parallel computing. 
In this paper, based on the theory of self-attention mechanism and state-of-the-art realization of self-attention in language models, we propose a general scheduling algorithm, which is derived from the optimum scheduling for small instances solved by a satisfiability checking(SAT) solver, to parallelize typical computations of self-attention. 
Strategies for further optimization on skipping redundant computations are put forward as well, with which reductions of almost 25\% and 50\% of the original computations are respectively achieved for two widely-adopted application schemes of self-attention. 
With the proposed optimization adopted, we have correspondingly come up with another two scheduling algorithms. 
The proposed algorithms are applicable regardless of problem sizes, as long as the number of input vectors is divisible to the number of computing units available in the architecture. 
Due to the complexity of proving the correctness of the algorithms mathematically for general cases, we have conducted experiments to reveal their validity, together with the superior quality of the solutions provided by which, by solving SAT problems for particular instances. 
\end{abstract}

\begin{IEEEkeywords}
mapping, data-flow, parallel computing, self-attention, SAT problem
\end{IEEEkeywords}
\vspace{-1em}
\section{INTRODUCTION\vspace{-0.5em}}
Remarkable achievements in the research field of deep learning have been realized during the past few years. 
As one of them, proposition of the concept of self-attention\cite{b1} is proved to be a remarkable success in the domain of natural language processing(NLP), since it successfully reveals the relation between any two words when conducting semantic analysis on a sentence, regardless the distance between them. 
Recent researches on self-attention, attracting attentions from the research community, includes but not limited to: 
further improving the theory of self-attention mechanism, for a better performance in certain implementations, like speech recognition\cite{b2}; 
establishing models with layers of self-attention applied, realizing their cooperative work with mature approaches, like the well-known convolutional modules\cite{b3}\cite{b4}.

However, together with significant improvements in accuracy, larger data set and bigger models nowadays are bringing about increasing requirement of computing resources as well, making computation power a bottleneck. 
Therefore, to speed up the computing process, there is an obvious trend to make use of different kinds of many-core machines as accelerators, in most of the current implementations of deep learning. 
In consequence, the significance of generating parallel scheduling solutions for the target implementation is beyond question, in order to load the corresponding computations onto hardware to conduct them efficiently. 
Nevertheless, without the guidance from a scheduling algorithm, there exist endless number of potential calculation orders related to a certain implementation of large scale, which have made the process of compilation extremely time consuming, or even unfeasible. 

To address this issue, there have been impressive attempts in the literature: 
Chin et al, have managed to establish the data flow graph with a framework from the level of assembly language\cite{b5}. 
But it is obviously not user-friendly and we believe that a high-level approach, which defines the behaviors of hardware resources in each computation cycle, can dig out more opportunities of potential parallelization, as the order of operations are fixed in assembly languages; 
Liu et al, have creatively applied the technology of reinforcement learning(RL) to obtain scheduling solutions, while the implementation turns out to suffer from weak scalability\cite{b6}. 
Miyasaka et al, have proposed remarkable results on the parallel scheduling of the computation of Matrix-Vector Multiplication(MVM)\cite{b7}. 
Although the results cannot be applied directly in our work, since our target of self-attention mechanism consisting of phases of computations more than just MVM, the methodology of interpreting the scheduling problems into incremental SAT problems has inspired us. 

In this paper, with no intention to propose any improvement on the theory of self-attention mechanism, we are aiming at optimizing the computations of self-attention by skipping the redundant ones that make no difference on the result of the applications. 
We propose general algorithms to efficiently schedule the optimized data-flow onto processing engines(PEs) connected as a unidirectional ring architecture. 
To the best of our knowledge, this is the first try on the realization of parallel computing targeting self-attention mechanism. 
In experimental phases, we interpret a parallel scheduling problem into a satisfiability checking(SAT) problem with high-level description, each variable in which decides a certain action of a PE. 

Based on two widely adopted application schemes of self-attention in practice, we have successfully:
1) respectively removed near 25\% and 50\% of the computations involved in the original definition of self-attention mechanism;
2) put forward general scheduling algorithms, targeting both of the computations with and without adoption of the proposed optimization on skipping avoidable calculations; 
3) proved the effectiveness and correctness of the proposed algorithms through solving SAT problems for specific cases. 

The paper is organized as follows. 
Section 2 gives the definition of the adopted parallel computing structure, and explains how to solve a parallel scheduling problem as a SAT problem. 
Section 3 summarizes typical computations of self-attention, and proposes an algorithm to schedule them onto target hardware architecture. 
Section 4 and 5 clarify our ideas of skipping unnecessary computations for two widely-received implementations of self-attention, with corresponding algorithms for parallel scheduling put forward. 
Section 6 demonstrates experimental results, serving as proofs of correctness and evaluations on the performance of the proposed algorithms. 
Section 7 concludes the paper and highlights our future work. 
\vspace{-0.5em}

\section{PARALLEL COMPUTING STRUCTURE\vspace{-0.5em}} 
\subsection{Adopted Ring-Connected Architecture\vspace{-0.5em}} 
Recent implementations of deep learning heavily rely on hardware as accelerators for computing resources. 
With those hardware devices adopted, such as GPUs, FPGAs and various types of ASICs, high performance computing can be realized by well schedule the computations of the target implementations onto the computing resources, which notably reduces the time cost of implementing deep learning techniques. 

Common communication structures among PEs includes mesh, torus, and even more complicated network-on-chip(NOC). 
In this work, we focus on parallel scheduling the data-flow onto a unidirectional ring-connected architecture\cite{b8}, which is a common definition adopted in many practical systems and can be smoothly realized as hardware\cite{b9}. 
\vspace{-1.5em}
\begin{figure}[htbp]
\centering
\subfloat[Relation between ring and mesh architectures]{
	\includegraphics[width=160pt]{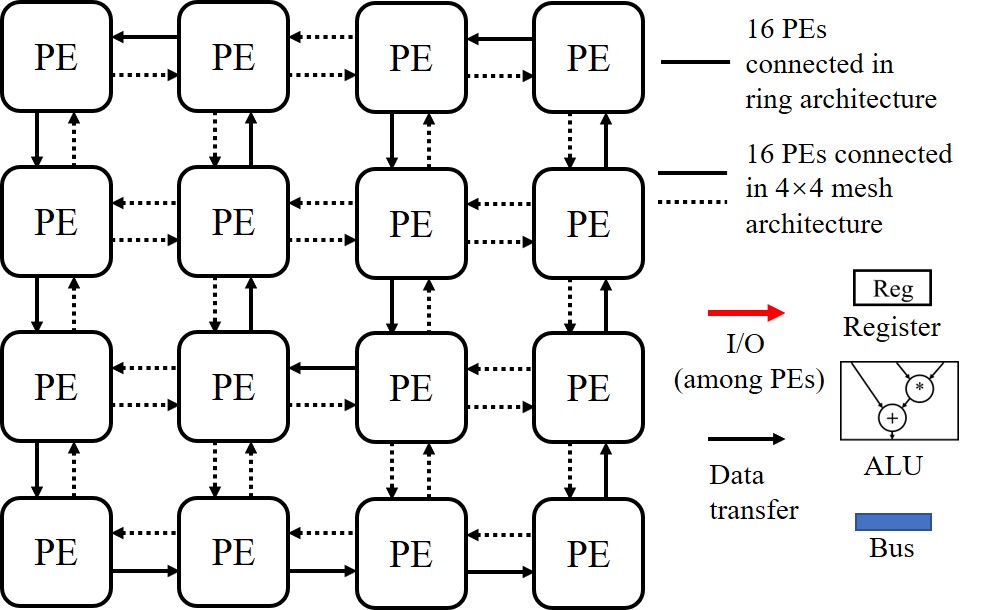}
	\label{subfig:Ring and mesh}
}
\hspace{-5pt}
\subfloat[Data path in each PE]{
	\includegraphics[width=80pt]{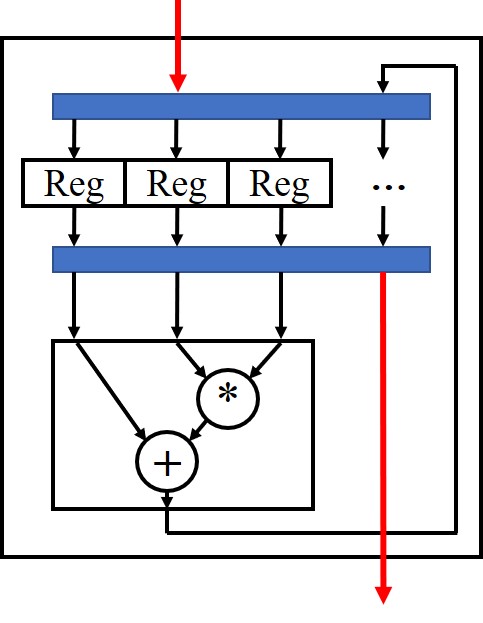}
	\label{subfig:Data path}
}
\vspace{-0.5em}
\caption{The adopted ring-connected architecture}
\label{fig:Ring-connected architecture}
\vspace{-1.5em}
\end{figure}

The reasons we are preferring ring, are: 
on the one hand, connections among PEs in a ring are simple enough. 
In general, if less hardware resources including communication is required, it is definitely better from the realization viewpoint; 
on the other hand, ring architecture is inherently a reduced version of others(as Fig. \ref{subfig:Ring and mesh} shows), which means as long as a scheduling solution achieving satisfying processing throughput is available for a ring, extra data path could be regarded as redundant, and the solutions for the former could be easily reused by the latter. 

Each PE is equipped with a set of arithmetic and logic units(ALUs) and registers, the numbers of which are specified by users. 
Operation flow of a PE in each computation cycle follows the order of \textit{\{multiplication/exponent operation/division\}-\{accumulation\}-\{data transfer\}}, while operations in all PEs are regarded as synchronous. 
This can be illustrated as a data path, and Fig. \ref{subfig:Data path} is giving an example, while all kinds of above-mentioned arithmetic operations besides multiplication are processible by the ALUs.  
Therefore, a scheduling solution is essentially determining the specific action of each PE, including: which kind of arithmetic operation to be conducted; which data to be processed as operands; to which location the computed data is located. 
We express such mapping problems, covering communication constraints among PEs from the architecture, together with target computations, as SAT problems. 
\vspace{-0.5em}

\subsection{Corresponding SAT Problem\vspace{-0.5em}}
\label{subsec:Corresponding SAT Problem}
The process of solving a SAT problem is essentially figuring out whether there is an assignment of variables leading the involved logic formula to be $1$, which is automatically taken care of by SAT solvers. 
For the convenience of comprehension, we demonstrate how to interpret an action of data transfer as an example:  
Assume the variable of $a(t_0,m_0)$ stands for the situation of ``data $a$ is in the local memory of PE-$m_0$ in cycle $t_0$'', while $a_{trans}(t_0,m_0)$ implies that ``in cycle $t_0$, PE-$m_0$ is sending the data of $a$ to its adjacent PE, PE-$(m_0+1)$. 
Considering the architecture of a ring, we could easily derive the four relations below: 

1) $a_{trans}(t_0,m_0)\  \rightarrow \ a(t_0,m_0)\  \wedge \ a(t_0+1,m_0+1)$. 
If the data of $a$ is transferred by PE-$m_0$ in cycle $t_0$, then it must be stored in PE-$m_0$ before being sent, and available in PE-$(m_0+1)$ after the phase of data communication of cycle $t_0$. 

2) $\overline{a_{trans}(t_0,m_0)}\  \rightarrow \ \overline{a(t_0,m_0)\  \wedge \ a(t_0+1,m_0+1)}$. If PE-$m_0$ does not transfer the data of $a$ in cycle $t_0$, then the situation mentioned in 1) cannot happen. 

3) $\overline{a(t_0,m_0)}\  \rightarrow \  \overline{a_{trans}(t_0,m_0)}$. If $a$ does not exist in PE-$m_0$ in cycle $t_0$, the mentioned data transfer could never occur. 

4) $\overline{a(t_0+1,m_0+1)}\  \rightarrow \  \overline{a_{trans}(t_0,m_0)}$. Ditto.  

If there is an assignment of variables that meets all the mentioned relations, the problem is proved to be satisfiable. 
Similarly, an assignment of variables, which satisfies the complete formula describing the whole scheduling problem, uniquely corresponds to a scheduling solution of target implementation. 
\vspace{-0.5em}

\section{SCHEDULING SELF-ATTENTION MECHANISM\vspace{-0.5em}}
\subsection{Typical Computation of Self-Attention\vspace{-0.5em}}
The core computations of self-attention can be summarized into the following three phases: 
\vspace{-1.5em}
\begin{sequation}
phase\ 1: \ w'_{ij} = \bm{q_i^T}\cdot \bm{k_j} = \sum_{l=1}^d \bm{q_i}[l]\cdot \bm{k_j}[l]
\end{sequation}
\vspace{-1em}
\begin{sequation}
phase\ 2: \ w_{ij} = softmax(w'_{ij}) = \frac{exp(w'_{ij})}{\sum_{j=1}^n exp(w'_{ij})}
\end{sequation}
\vspace{-1em}
\begin{sequation}
phase\ 3: \ \bm{y_i} = \sum_{j=1}^d w_{ij}\cdot \bm{v_j}
\vspace{-1.5em}
\end{sequation}

It is clear that the $n$ $d$-dimensional outputs, \textbf{y}, is a weighted sum of value vectors(\textit{phase3}), \textbf{v}, while the $n^2$ weights are determined by the matching degree of query and key vectors, \textbf{q} and \textbf{k}, numerically decided by a dot product(\textit{phase1}), with softmax operations straight after for normalization (\textit{phase2}). 

The vectors of \textbf{q}, \textbf{k} and \textbf{v} are obtained by conducting linear projections on the $n$ $d$-dimensional inputs, \textbf{x}, with three $d\times d$ square matrices of $Q$, $K$ and $V$, which are prepared in advance by learning from the data sets of users' specific implementation. 
From the perspective of a language model, each input can be interpreted as a sentence containing $d$ words. 

The pre-process of projecting \textbf{x} to be \textbf{q}, \textbf{k} and \textbf{v}, is essentially typical MVM calculations, the scheduling solutions of which can be easily obtained by a method, e.g. the one shown in\cite{b10}. Thus, in the ensuing discussion, we directly regard the three intermediate vectors, rather than \textbf{x}, as input data. 

Besides, there are usually other tricks applied, such as \textit{multi-head} mechanism for addressing the ambiguity of natural language. 
However, due to their little impact on the core computations determined by the mentioned three phases, they are not explicitly discussed in this paper, for simplicity. 
For example, the concept \textit{multi-head} refers to multiple sets of $Q$, $K$ and $V$ matrices, while operations for each \textit{head} are the same as the summarized three phases. 
Therefore, our contribution is applicable for the implementations adopting those tricks. 

As far as we know, application schemes for the mechanism of query, key and value vary in existing works: 
using the same matrix as both $K$ and $V$, which is inherited from previous research on implementing attention mechanism for translation\cite{b11}; 
utilizing one vector for all the three roles is another emerging scheme that is well-received\cite{b1}. 
For some schemes, if operands in two phases of computations are the same, with a well-scheduled data movement, a reuse of the inputs remained in the registers of PEs after the former phase is possible. 
Thus, there is an opportunity to minimize the requirements of data transfer between PEs and external memory system(where input data are originally stored), which is recognized as extremely time consuming. 
To be clear, the load/storage of input/output data from/to the external memory is not our concern, since we are focusing on completing the whole computation process with only necessary data communication among PEs. 
Therefore, we assume the input data required for computations are already loaded into PEs before computation begins. 
 
From the above, we take into consideration two widely used application schemes of self-attention mechanism, where: 1) the same vectors are simultaneously adopted as query, key and value, which is the most interesting but complicated situation, since the computations are theoretically possible to be completed with loading the $n$ input vectors only once, and reuse them for all the phases; 2) the three roles are played by different projections from the input vectors, which is straightforward but guarantees our work to be without loss of generality. 

However, instead of the mentioned specific application schemes, discussion begins with the original definition of self-attention mechanism, serving as the baseline for optimization.
\vspace{-2em}

\subsection{Baseline of Scheduling Solution\vspace{-0.5em}}
\label{subsec:Baseline of Scheduling Solution}
To avoid excessive subjectivity, we regard the scheduling problem targeting the realization of self-attention mechanism in \cite{b12}, which is recognized as a widely used and state-of-the-art deep learning framework, as the baseline. 

The realization includes: \textit{phase1}: $dn^2$ times of multiply-accumulation-computations(MACs), since production of each of the $n^2$ weights involves $d$ times of MACs; 
\textit{phase2}: $n^2$ times of exponent operation-accumulation-computations(EACs), together with $n^2$ times of division, determined by the definition of softmax operation; 
\textit{phase3}: another $dn^2$ times of MACs, every $dn$ times to produce one of the $n$ output vectors of \textbf{y}. 

We format computations in \textit{phase1, 3} into MVM, with the assumption that $n$ is equal to $d$, which ensures the involved matrices to be square, while adopting $m$, the number of available PEs in the parallel structure, to be divisible by $n$. 
The mentioned assumptions are proved to be important premises for the existence of a regular scheduling solution for MVM, according to \cite{b10}. 
An algorithm to schedule the target computations is inductively derived from the solution instances given by solving corresponding SAT problems of small sizes. 
\vspace{-1.4em}
\begin{algorithm}
  \caption{General scheduling algorithm to parallelize computations for original definition of self-attention}
  \tiny
  \label{alg:General scheduling algorithm for baseline}
  \KwIn{$n$ $n$-dimensional vectors of \textbf{q}(query), \textbf{k}(key), and \textbf{v}(value)}, respectively
  \KwOut{$n$ $n$-dimensional vectors of \textbf{y}}
  // phase1: \\
  \For{$h\in [1,n]$}
  {
    \For{$i\in [1,\frac{n}{m}]$}
    {
      \For{$j\in [1,n]$}
      {
      	\For{$l\in [1,m]$}
      	{
      	  $A = h$; $B = ((l-j)\%m+(i-1)\cdot m-1)\%n+1$; \\  
          $C = (l+(j-1)\backslash m\cdot m+(i-1)\cdot m-1)\%n+1$; \\
      	  $w'_{AB} += \bm{q_A}[C]\cdot \bm{k_B}[C]$; \\
      	  \If{$j\neq n$}
      	  {
      	    transfer $w'_{AB}$ to the adjacent PE; \\
      	  }      	    
      	}
      }
    }
  }
  // phase2 - step1: conduct EAC. \\    
  \For{$i\in [1,\frac{n}{m}]$}
  {
    \For{$j\in [1,n]$}
    {
      \For{$l\in [1,m]$}
      {
        $A = (l-j)\%m+(i-1)\cdot m+1$ \\
        $B = (l+(j-1)\backslash m\cdot m+(i-1)\cdot m-1)\%n+1$ \\
        $sum_A += exp(w'_{AB})$; transfer $sum_A$ to the adjacent PE; \\
      }
    }
  }
  // phase2 - step2: conduct division. \\ 
  $\quad$// loops similar to phase2 - step1, with local computation of $w_{AB} = exp(w'_{AB})/sum_A$, \\
  $\quad$computations are completed by always sending local data of $sum_A$. \\
  // phase3: \\
  $\quad$// loops similar to phase1, with local computation of $\bm{y_A}[C] += w_{AB}\cdot \bm{v_B}[C]$, while $B = (l-j)\%m+(i-1)\cdot m+1$. \\
  $\quad$Computations are completed by always sending local data of $w_{AB}$. \\
\end{algorithm}
\vspace{-1.6em}

Since the $m$ PEs are working simultaneously in each computation cycle, the most inner loops are conducted in parallel. 

\begin{figure}[htbp]
\centering
\includegraphics[width=200pt]{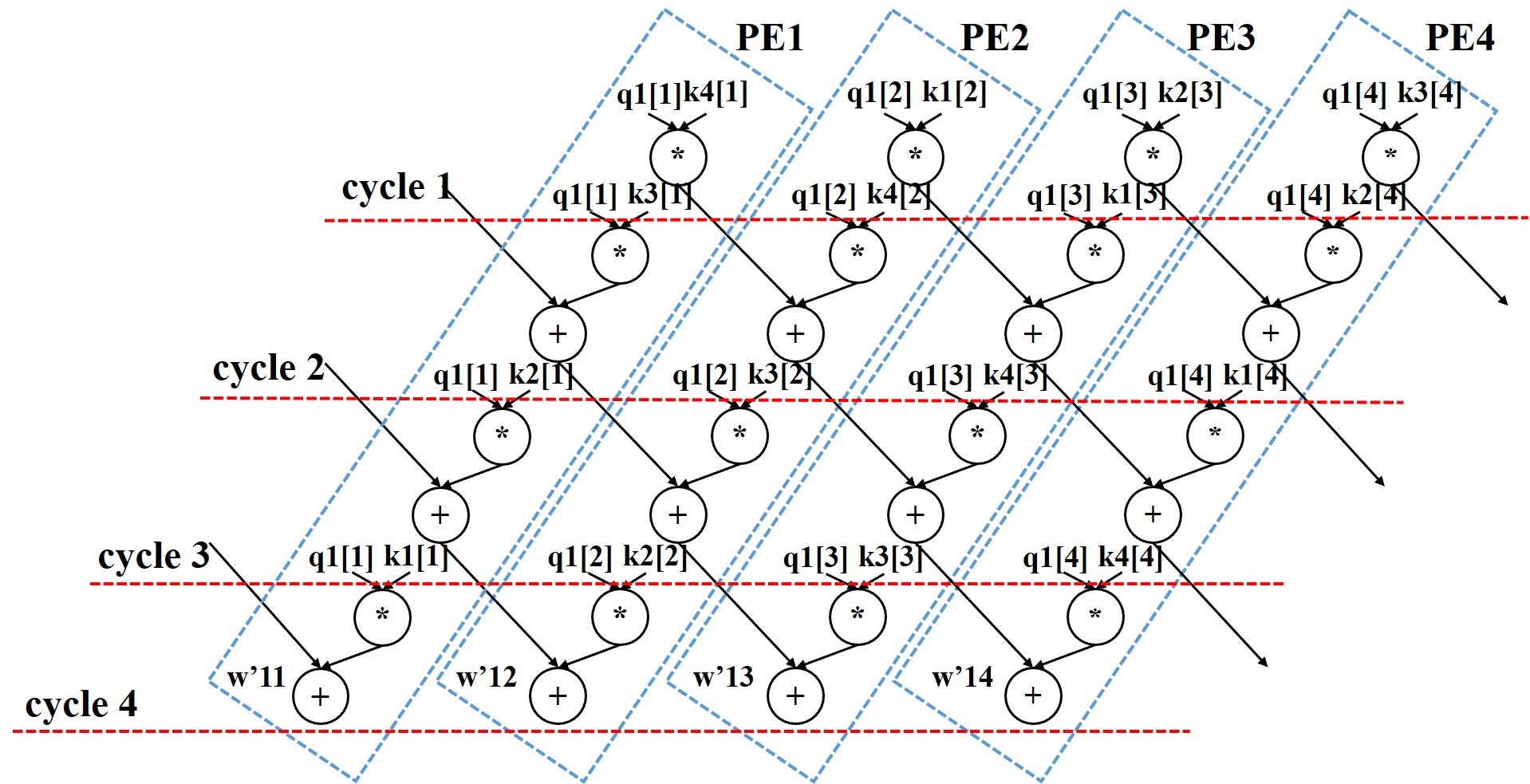}
\vspace{-0.8em}
\caption{Partial scheduling solution to computations of \textit{phase1}, described by our general algorithm}
\label{fig:Partial scheduling solution for original dataflow}
\vspace{-1.4em}
\end{figure}

For the ease of comprehension, the partial scheduling solution of the MACs conducted in \textit{phase1} in the case of $n$=$d$=$m$=$4$ is shown in Fig. \ref{fig:Partial scheduling solution for original dataflow}. 
It takes 7.49 seconds for the solver to find this solution from the corresponding SAT problem, which contains 330 thousand variables. 
As a contrast, Algo.\ref{alg:General scheduling algorithm for baseline} is providing a scheduling solution without any time cost, while a further verification of which costs only 0.05 second.   
\vspace{-1.2em}

As shown in Algo.1, all the PEs are associated with arithmetic operations during the whole computing process, we can state that an optimum usage of computing resources is achieved by our algorithm, without any extra data communication between external memory and PEs to reorder the data distribution. 

To support the computations, there is a balanced requirement of local memory for each PE to store O($n^2/m$) elements of data, which is easy to be realized from the perspective of hardware manufacturing, because of the uniformity. 

\section{OPTIMIZATION WHEN QUERY, KEY AND VALUE ARE THE SAME VECTORS\vspace{-0.5em}}
\label{sec:Optimization when query key and value are the same}
\subsection{Redundant Computations: Symmetric Weights Matrix\vspace{-0.5em}}
\label{subsec:Cause of redundant computations when query key and value are the same}
For the implementations where the vectors of query, key and value are adopted to be the same, $dn^2$ times of MACs in \textit{phase1} is proved to be a waste of computing resources, since for any weights of $w'_{ij}(i,j \in [1,n])$, $w'_{ij}$ is numerically equal to $w'_{ji}$:
\vspace{-0.6em}
\begin{sequation}
w'_{ij} = \bm{x_i^T}\cdot \bm{x_j} = \sum_{l=1}^d \bm{x_i}[l]\cdot \bm{x_j}[l] = w'_{ji}
\vspace{-0.6em}
\end{sequation}
Please notice that although still represented as \textbf{x} here for simplicity, the vectors of \textbf{x} are likely to be the linear mapped ones(decided by users), rather than the original input vectors. 

That is to say, to remove redundant computations, we should calculate only one element in each pair of ($w'_{ij}$, $w'_{ji}$), and reuse it as the the other one in following phases of computations. 
In other words, in the $n\times n$ weight matrix, only computations related to the $n$ main diagonal elements and half of the off-diagonal elements are necessary to be covered. 
Please notice that the selection of the off-diagonal elements to be calculated is operated simultaneously in the two triangles of the weight matrix, not limited to elements belonging to the same triangle, as the only target is to minimize the need of extra data transfer to reuse the calculated data as their symmetric ones.

At the same time, since the equivalence of $w'_{ij}$ and $w'_{ji}$ revealed in Eq.(4) is no longer valid after softmax operation, the recognized opportunity is restricted to \textit{phase1}, i.e. reducing the MACs in \textit{phase1} from $dn^2$ to $dn(n+1)/2$ times. 
Computations in the following two phases remain the same, for which we therefore reuse the corresponding part in Algo.1. 
\vspace{-0.5em}

\subsection{Scheduling Algorithm Utilizing the Symmetry\vspace{-0.5em}}
\label{subsec:Scheduling algorithm utilizing the symmetry} 
To come up with a scheduling algorithm consistent with the proposed scheme to reduce number of computations, the methodology on selecting the weights to be calculated is the first to be decided. 
\vspace{-1.1em}
\begin{figure}[htbp]
\centering
\subfloat[$n$ is odd(e.g. $n=5$)]{
	\includegraphics[width=120pt]{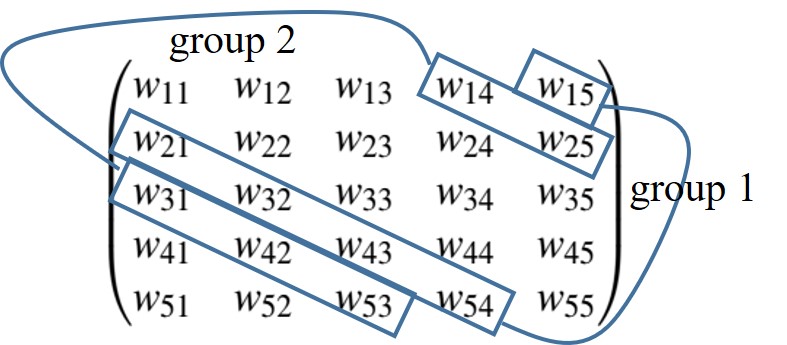}
	\label{subfig:Packing:n is an odd number}
}
\hspace{5pt}
\subfloat[$n$ is even(e.g. $n=4$)]{
	\includegraphics[width=95pt]{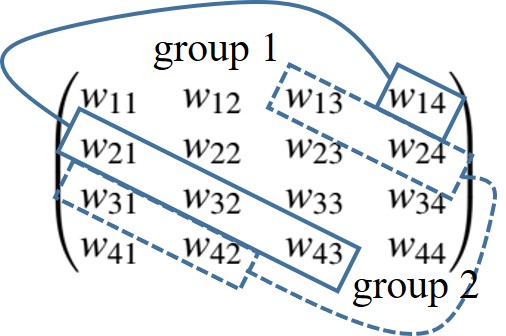}
	\label{subfig:Packing:n is an even number}
}
\vspace{-0.7em}
\caption{Packing selected elements into groups}
\label{fig:Packing selected elements into groups}
\vspace{-0.7em}
\end{figure}

\vspace{-0.8em}
Based on SAT solutions for small instances, we propose the following scheme for selection, with examples shown in Fig.\ref{fig:Packing selected elements into groups}: 

1) Since data $w'_{ij}$ is supposed to locate in the $(j\%m)$-th PE, when \textit{phase1} finishes(Algo.1), diagonal is decided to be the basic unit of selection, since elements in the same diagonal requires same times($m\%(i-j)$) of transfer for data reuse. 

2) For diagonals in the upper triangular part, the selection begins from the one most distant from the main diagonal, while selecting from the one closest to the main diagonal for lower triangular diagonals of elements, since their data reuse requires minimum numbers of extra data transfer. 

3) Regard $n$ elements belonging to the $i$-th selected upper triangular diagonal and the $i$-th selected lower triangular diagonal as a group(marked with a curve in Fig.\ref{fig:Packing selected elements into groups}), $i\in [1,(n-1)/2]$ if $n$ is odd, $i\in [1,n/2-1]$ if $n$ is even. 
Schedule the computations related with elements from the same group equivalently.  

4) When $n$ is even, $n$ elements in the $(n/2)$-th upper triangular and lower triangular diagonals are symmetric to each other(marked with dotted squares in Fig.\ref{subfig:Packing:n is an even number}). 
Therefore, to avoid redundant computations, only $n/2$ numerically different ones should be calculated. 
However, considering these $n/2$ elements cannot be packed with any other elements into a group, leading to an unbalanced scheduling problem, we apply the trade off to cover $dn/2$ redundant MACs. 
In other words, all of these $n$ weights squared in dotted lines are forming another group and to be calculated. 
This trade off is unnecessary when $n$ is odd, since all the $n(n-1)/2$ off-diagonal elements to be calculated can be evenly divided into $(n-1)/2$ groups. 

Following the introduced methodology, we propose another algorithm, Algo.2, to schedule only the reduced computations. 
\vspace{-2em}
\begin{algorithm}
  \tiny
  \caption{General scheduling algorithm to parallelize reduced computations utilizing symmetry of weights}
  \label{alg:General scheduling algorithm for optimization 1}
  \KwIn{$n$ $n$-dimensional vectors of \textbf{x}}
  \KwOut{$n$ $n$-dimensional vectors of \textbf{y}}
  // phase1: produce $n^2$ intermediate weights, $w'$, $(n(n+1)/2)$ of which are calculated, while the others are obtained by data reuse. \\
  $h' = (n\ is\  odd)?(n-1)/2:(n/2-1)$; \\
  // $h'$ represents number of packed groups.
  \For{$h\in [0,h']$}
  {
    \For{$i\in [1,\frac{n}{m}]$}
    {
      \For{$j\in [1,n]$}
      {
      	\For{$l\in [1,m]$}
      	{
      	  $B = ((l-j)\%m+(i-1)\cdot m-1)\%n+1$; \\
      	  $C = (l+(j-1)\backslash m\cdot m+(i-1)\cdot m-1)\%n+1$; \\
          \If{$(h\%m)\leq (m/2)$} 
          {         
            $A = (B+h-1)\%n+1$; 
          }
          \Else
          {
            $A = (B-h-1)\%n+1$; 
          }
      	  $w'_{AB} += \bm{x_A}[C]\cdot \bm{x_B}[C]$; \\
      	  \If{$j=n$}
      	  {
      	    \If{$h\geq 1$}
      	    {
      	      transfer the data of $w'_{AB}$ to the adjacent PE; \\
      	      \If{$(h\%m)=1$}
      	      {
      	        store $w'_{AB}$ as $w'_{BA}$ in the adjacent PE; \\ 
      	      }
      	    }
      	  }
      	  \Else
      	  {
      	    transfer the data of $w'_{AB}$ to the adjacent PE; \\	
          }
      	}
      }
      // for off-main diagonal elements, conduct extra data transfer to realize data reuse. \\
      \If{$(h\%m)>1$}
      {
        $p' = ((h\%m)\leq (m/2))?(h\%m):(m-h\%m)$; \\
        \For{$p\in [1,p']$}
        {
          $E = (l-p-1)\%m+(i-1)\cdot m +1$; \\
          \If{$(h\%m)\leq (m/2)$}
          {         
            $D = (E+h-1)\%n+1$;
          }
          \Else
          {
            $D = (E-h-1)\%n+1$;
          }
          \For{$l\in [1,m]$}
          {
            transfer $w'_{DE}$ to the adjacent PE; \\
            \If{$p=p'$}
            {
      	      store $w'_{DE}$ as $w'_{ED}$ in the adjacent PE;
      	    }
          }
        }
      }
    }
  }
  // if $n$ is even, apply the introduced trade off for a balanced work load. \\
  $\quad$// similar to the previous stage, but $A = (B+n/2-1)\%n+1$ and no need of extra data transfer for reuse. \\
  // phase2, 3: the same as phase2, 3 in Algo.1. \\ 
\end{algorithm}
\vspace{-1.5em}

As we have realized data reuse by operating data communication among PEs, loading/storing the input/output data, which is not under discussion of the proposed algorithm, is all the expensive data transfer between external memory and PEs needed for the whole process, which is no doubt the optimum. 

Besides, in the parallel scheduling solutions following Algo.2, only the necessary $dn(n+1)/2$(when $n$ is odd), or $dn(n+2)/2$(when $n$ is even), times of computations are involved in \textit{phase1}, out of the original $dn^2$ in Algo.1. 
Considering the total computations in all the three phases are $(2dn^2+2n^2)$,
we can state that Algo.2 has further achieved a save of near 25\% of computations.

\section{OPTIMIZATION FOR MASKED SELF-ATTENTION\vspace{-0.5em}}
\label{sec:Optimization for masked self-attention}
\subsection{Masked Self-attention\vspace{-0.5em}} 
\label{subsec:Masked Self-attention}
In implementations of NLP, self-attention operation in decoders has to be ensured that the production of any element in the output sequence is depending only on the already established ones\cite{b1}, which can be interpreted as an imitation of human activity: when making a sentence one word after another, the latter words make no difference to the former ones which are already decided. 
Accounting for this theory, a widely known trick called \textit{mask}, is introduced into the realization of self-attention in language models. 

The concept determines that, theoretically only the calculations related to the $n(n+1)/2$ elements, located on the main diagonal and the lower triangular part of the weight matrix are responsible for the production of output vectors, which is another opportunity to skip redundant computations. 

We again refer to \cite{b12} for the realization of masked self-attention in practice. 
It is recognized that in current implementations, the effect of mask is realized by assigning the masked weights to be negative infinity to eliminate their impact on the output vectors, after the weight matrix is produced in \textit{phase1}. 
That is to say, there does exist a room for removing redundant computations when producing the weight matrix. 

It should be noticed that, since those masked weights can be excluded from all the three phases of computations, redundant computations spread over the whole computing process, rather than only \textit{phase1} as the former implementation in section \ref{sec:Optimization when query key and value are the same}. 
\vspace{-0.5em}

\subsection{Scheduling Algorithm Considering Masked Self-attention\vspace{-0.5em}} 
\label{subsec:Scheduling algorithm considering masked self-attention}
From the experience of proposing Algo.2, we easily come up with another strategy for the current implementation, on packing the elements to be computed into groups for scheduling: 
Through trials on small instances, we adopt rows as the basic units, which can guarantee a high processing throughput in \textit{phase3}, instead of diagonals. 
In the current circumstance, all the elements to be calculated are located in lower triangle and main diagonal of the weight matrix. 
To ensure the number of elements in each group is $n$, for a balanced workload, we pack the $i$-th and $(n-i)$-th rows as group-$i$. 
In this way, optimal scheduling involving only the minimum computations proves to be achievable for \textit{phase1,3}, without extra data communication between external memory system and PEs for reordering. 

However, the real problem is that different types of data distribution of weights are required for optimal processing throughput in \textit{phase2} and $3$, pointing out the lack of a general scheduling solution simultaneously satisfying needs of all the three phases of computations, explained by Fig.4.
\begin{figure}[htbp]
\centering
\subfloat[Distributed-type(\textit{phase3})]{
	\includegraphics[width=90pt]{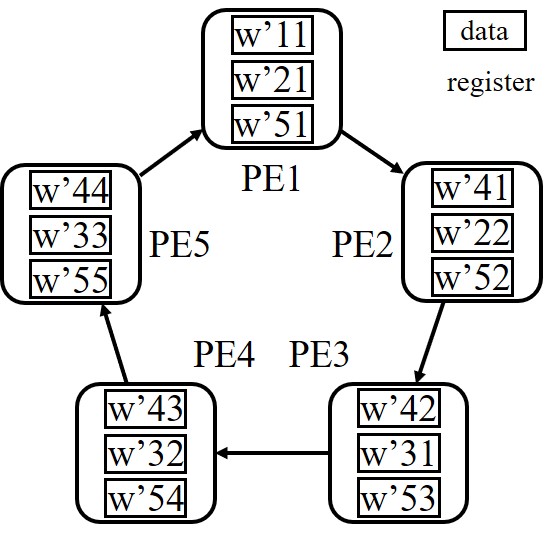}
	\label{subfig:Distributed-type}
}
\hspace{5pt}
\subfloat[Centralized-type(\textit{phase2})]{
	\includegraphics[width=90pt]{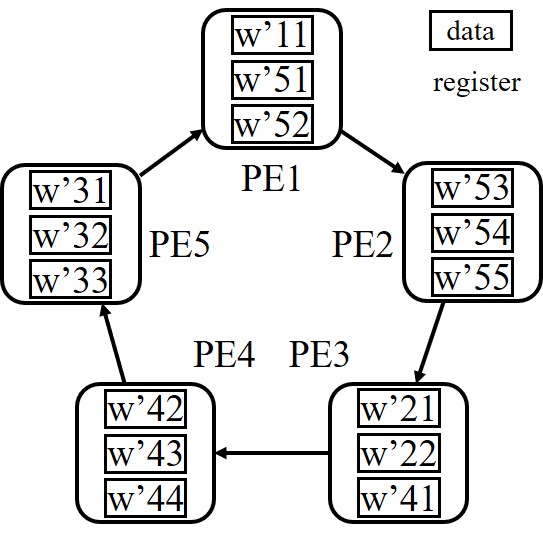}
	\label{subfig:Centralized-type}
}
\vspace{-0.5em}
\caption{Types of weights' distribution required in different phases for high parallelization(e.g. $n=5$)}
\label{fig:Types of data distribution}
\vspace{-0.5em}
\end{figure}

Here are our definitions for this two kinds of data distribution: 
1) Distributed-type. For all the $i$ elements in the $i$-th row of the lower triangle and main diagonal of the weight matrix, they are distributed in all the $m$ PEs, with $\lfloor i/m \rfloor$ or $\lceil i/m \rceil$ in each PE;
2) Centralized-type. The above-mentioned $i$ elements are concentrated in only minimum number($\lceil \frac{2im}{n(n+1)} \rceil$) of PEs. 

Based on our experiments, there is no regular way to convert the type of weights' distribution from centralized one(required in \textit{phase2}) to distributed one (required in \textit{phase3}) during the computations of \textit{phase2}. 
Therefore, in our proposed algorithm utilizing the feature of masked self-attention, we apply the trade off to allow a near 50\%($\frac{n(n+1)/2}{n\cdot n}$) hardware utilization rate during \textit{phase2}, for optimal processing throughput in \textit{phase3}, as more computations are involved in the latter. 
\vspace{-1.2em}
\begin{algorithm}
  \caption{General scheduling algorithm to parallelize reduced computations considering masked self-attention}
  \tiny
  \label{alg:General scheduling algorithm for optimization2}
  \KwIn{$n$ $n$-dimensional vectors of \textbf{q}(query), \textbf{k}(key), and \textbf{v}(value), respectively}
  \KwOut{$n$ $n$-dimensional vectors of \textbf{y}}
  // phase1: \\
  $h' = (n\ is\  odd)?(n-1)/2:(n/2-1)$; \\
  \For{$h\in [0,h']$}
  {
    \For{$i\in [1,\frac{n}{m}]$}
    {
      \For{$j\in [1,n]$}
      {
      	\For{$l\in [1,m]$}
      	{
      	  $A = n-h$; $B = (l-j)\%m+(i-1)\cdot m+1$ \\
      	  $C = (l+(j-1)\backslash m\cdot m+(i-1)\cdot m-1)\%n+1$; \\       
      	  \If{$A<B$}
      	  {  
      	    $A = h$; $B \%= (n-h)$; \\
      	  }
      	  $w'_{AB} += \bm{q_A}[C]\cdot \bm{k_B}[C]$; transfer $w'_{AB}$ to the adjacent PE;	
      	}
      }
    }
  }
  // if $n$ is even, for computations of the ($n/2$) elements on the ($n/2$)-th row: \\
  $\quad$// similar to the previous stage, computations are conducted only if $A\geq B$, but $A$ is always $(n/2)$. \\
  // phase2 - step1: conduct EAC. \\    
  \For{$i\in [1,\frac{n}{m}]$}
  {
    \For{$j\in [1,n]$}
    {
      \For{$l\in [1,m]$}
      {      
        $A = (l-j)\%m+(i-1)\cdot m$ \\
        $B = [l+(j-1)\backslash m\cdot m+(i-1)\cdot m-1]\%n+1$ \\
        \If{$A<n/2$}
        {
          $B =(B-(n-A)-1)\%n+1$;
        }        
        \If{$A\geq B$}
        {
          $sum_A += exp(w'_{AB})$; \\
        }        
      	transfer $sum_A$ to the adjacent PE;
      }
    }
  }
  // phase2 - step2: conduct division. \\ 
  $\quad$// loops similar to phase2 - step1, with local computation of $w_{AB} = exp(w'_{AB})/sum_A$. \\
  $\quad$Computations are completed by always sending local data of $sum_A$. \\
  // phase3: \\
  $\quad$// loops similar to phase1, with local computation of $\bm{y_A}[C] += w_{AB}\cdot \bm{v_B}[C]$. \\
  $\quad$Computations are completed by always sending local data of $w_{AB}$. \\
\end{algorithm}

\vspace{-1.4em}
According to the proposed Algo.\ref{alg:General scheduling algorithm for optimization2}, during the whole computing process, there is still no data communication required between external memory system and PEs. 

Moreover, utilizing the mechanism of masked self-attention, the scheduling solutions are covering only the minimum number of computations: the ones related with the $n(n+1)/2$ unmasked elements located in the main diagonal and lower triangular part of the weight matrix, stating that a save of near 50\% of computing resources is achieved. 
\vspace{-0.5em}

\section{EXPERIMENT\vspace{-0.5em}}
Accounting for the complexity of mathematically proving the correctness of the proposed scheduling algorithms, experiments are conducted for the verification on some particular instances. 
Meanwhile, a direct comparison to reveal the superior quality of the solutions according to the algorithms, especially the ones following Algo.\ref{alg:General scheduling algorithm for optimization 1} and \ref{alg:General scheduling algorithm for optimization2}, which respectively utilize the features of the two introduced application schemes to avoid redundant computations, is another purpose of the experiments.

Exhaustive search-based method(marked as \textit{ES}) is adopted as the peer for comparison, with which the elements to be calculated in \textit{phase1} of the first implementation, and data transfer in \textit{phase2} of the second implementation are decided by exploring the search space exhaustively. 

It is important to notice that, as the proposed algorithms themselves are the scheduling solutions, solving corresponding SAT problems is essentially verifying whether they are meeting with the defined constraints, rather than looking for a solution under those constraints as \textit{ES} does. 
In other words, for the proposed algorithms, there is in strict sense no time cost required for generation of scheduling solutions.

All the experiments were conducted with a machine having 512GB RAM, and dual Intel Xeon E5-2699 v4 operating at 2.20 GHz. 
ABC\cite{b13} is adopted as the SAT solver, while one hour is set as the time limitation. 
\vspace{-0.6em}
\begin{table}[htbp]
\caption{SCHEDULING FOR IMPLEMENTATIONS WHERE QUERY, KEY, VALUE ARE THE SAME}
\vspace{-1.7em}
\begin{center}
\begin{tabular}{|c|c|c|c|c|c|c|c|}
\hline
\multirow{2}*{Size} & \multirow{2}*{\#PEs} & \multicolumn{3}{|c|}{Run time in sec} & \multicolumn{3}{|c|}{\#Cycles} \\
\cline{3-8} 
 &  & Algo.1 & \textit{ES} & Algo.2 & Algo.1 & \textit{ES} & Algo.2 \\
\hline
3 & 3 & 0.01 & 0.31 & 0.01 & 24 & 21 & 21 \\
\hline
4 & 4 & 0.05 & 2.73 & 0.04 & 40 & 35 & 36 \\
\hline
5 & 5 & 0.08 & 14.27 & 0.06 & 60 & 50 & 50 \\
\hline
6 & 3 & 0.37 & TO & 0.26 & 168 & 142$^*$ & 146 \\
\hline
6 & 6 & 0.19 & TO & 0.15 & 84 & 71$^*$ & 73 \\
\hline
\multicolumn{8}{|c|}{$\cdots$} \\
\hline
15 & 5 & 27.38 & TO & 14.45 & 1440 & 1134$^*$ & 1134 \\
\hline
15 & 15 & 17.32 & TO & 10.41 & 480 & 396$^*$ & 396 \\
\hline
\multicolumn{8}{l}{$^*$Data marked with asterisks are derived from (\#computations)/(\#PEs), } \\
\multicolumn{8}{l}{which is not likely to achieve in practice, especially for a general solution.}
\end{tabular}
\label{tab:For implementations where query, key, value are the same}
\end{center}
\vspace{-1.2em}
\end{table}

\vspace{-1em}
Based on results in Table \ref{tab:For implementations where query, key, value are the same}, \textit{ES} is not able to generate parallel scheduling solutions within the time limitation when $n\geq 6$, while both of the proposed algorithms are proved reliable. 

On the other hand, through utilizing the symmetry of weight matrix in the implementations where the three roles of query, key, value are played by the same vectors, Algo.\ref{alg:General scheduling algorithm for optimization 1} excludes all(except $n/2$ weights, when $n$ is even) the redundant computations, consistent with the derivation in section \ref{subsec:Scheduling algorithm utilizing the symmetry}. 
\vspace{-0.6em}
\begin{table}[htbp]
\caption{SCHEDULING FOR IMPLEMENTATIONS OF MASKED SELF-ATTENTION}
\vspace{-1.7em}
\begin{center}
\begin{tabular}{|c|c|c|c|c|c|c|c|}
\hline
\multirow{2}*{Size} & \multirow{2}*{\#PEs} & \multicolumn{3}{|c|}{Run time in sec} & \multicolumn{3}{|c|}{\#Cycles} \\
\cline{3-8} 
 &  & Algo.1 & \textit{ES}& Algo.3 & Algo.1 & \textit{ES}& Algo.3 \\
\hline
3 & 3 & 0.01 & 3.94 & 0.00 & 24 & 17(16)$^{\mathrm{a}}$ & 18 \\
\hline
4 & 4 & 0.05 & 27.40 & 0.01 & 40 & 26(25) & 32 \\
\hline
5 & 5 & 0.08 & TO & 0.04 & 60 & 36$^*$ & 40 \\
\hline
6 & 3 & 0.37 & TO & 0.11 & 168 & 84$^*$ & 120 \\
\hline
6 & 6 & 0.19 & TO & 0.11 & 84 & 42$^*$ & 60 \\
\hline
\multicolumn{8}{|c|}{$\cdots$} \\
\hline
15 & 5 & 27.38 & TO & 5.72 & 1440 & 768$^*$ & 810 \\
\hline
15 & 15 & 17.32 & TO & 5.29 & 480 & 256$^*$ & 270 \\
\hline
17 & 17 & TO & TO & 8.35 & 612 & 324$^*$ & 340 \\
\hline
\multicolumn{8}{l}{$^{\mathrm{a}}$Data between the parentheses are theoretical values proved to be} \\
\multicolumn{8}{l}{unachievable by SAT solver. Meaning of asterisks is the same as Table \ref{tab:For implementations where query, key, value are the same}.}
\end{tabular}
\label{tab:For implementations where query, key, value are different}
\end{center}
\vspace{-1.2em}
\end{table}

\vspace{-1em}
As introduced in section \ref{subsec:Scheduling algorithm considering masked self-attention}, due to the adopted trade off to accept a near 50\% processing throughput in \textit{phase2}, in exchange for optimal computing performance in \textit{phase3}, it seems that Algo.\ref{alg:General scheduling algorithm for optimization2} is inferior to the ones generated by \textit{ES} at the first glance. 
However, the theoretical numbers of cycles required by \textit{ES} cannot be achieved by practical solutions, since they are directly derived from (\#computations)/(\#PEs), without any consideration on constraints like data communication. 

More importantly, while \textit{ES} is still suffering from weak scalability in the current case of application scheme, Algo.\ref{alg:General scheduling algorithm for optimization2} has effectively removed near 50\% of the computations involved by Algo.\ref{alg:General scheduling algorithm for baseline}, which are related to those ``masked'' elements. 

Although the verification for correctness of the proposed algorithms is conducted only for cases with problem sizes up to 15(17) in the first(second) implementations, determined by the capability of a SAT solver, it should be clearly recognized that our general algorithms are applicable for implementations of any sizes. 
As an example of evidence, when the number of input vectors, $n$=10,000 and the number of PEs, $m$=5,000, Algo.\ref{alg:General scheduling algorithm for optimization2} is providing a parallel scheduling solution requiring $n^2(n+2)/m+2n^2/m$=200.08 million computation cycles.
\vspace{-0.5em}

\section{CONCLUSION\vspace{-0.5em}}
In this paper, we have addressed the problem of efficiently scheduling self-attention, a mechanism widely-adopted in the research field of NLP in recent years, onto parallel hardware, in order to realize parallel computing. 

Our contribution includes the proposition of: 
1) scheduling algorithm to realize parallel computing for typical computations of self-attention mechanism. 
2) reductions of near 25\% and 50\% of computations in 1), respectively for two well-received application schemes of self-attention; 
3) scheduling algorithms considering the optimization proposed in 2). 

Limited by SAT solvers' capability, we have verified the proposed algorithms for the cases with problem size, $n$, up to 15 and 17, respectively for the two involved application schemes. 
However, it is clear that the algorithms are with complete scalability and believed to be reliable, according to their proved correctness in all our trials on particular instances. 

To extend the verification for problems with larger sizes, using tools other than a SAT solver, e.g. a satisfiability modulo theories(SMT) solver as a theorem prover, like \cite{b10} did, has a place in our plans. 
In addition, to further extend compilers' circle of competence, general algorithms to schedule the computations of a complete deep learning model consisting of multiple layers and involving various mechanisms, are underlined in our future work as well. 


\vspace{-0.5em}

\begin{thebibliography}{00}
\vspace{-0.5em}
\scriptsize
\bibitem{b1} A. Vaswani, et al, ``Attention is All You Need,'' in \textit{Advances in Neural Information Processing Systems}, 2017, pp. 5998-–6008. 
\bibitem{b2} D. Povey, H. Hadian, P. Ghahremani, et al, ``A Time-Restricted Self-Attention Layer for ASR,'' in \textit{Proceedings of the IEEE International Conference on Acoustics, Speech and Signal Processing (ICASSP)}, 2018, pp. 5874--5878. 
\bibitem{b3} I. Bello, et al, ``Attention Augmented Convolutional Networks,'' in \textit{Proceedings of the IEEE International Conference on Computer Vision}, 2019, pp. 3286-–3295. 
\bibitem{b4} W. C. Kang and J. McAuley, ``Self-Attentive Sequential Recommendation,'' in \textit{Proceedings of the IEEE International Conference on Data Mining (ICDM)}, 2018, pp. 197–-206. 
\bibitem{b5} S. A. Chin, J. H. Anderson, ``An Architecture-Agnostic Integer Linear Programming Approach to CGRA Mapping,'' in \textit{Proceedings of the 55th Annual Design Automation Conference}, 2018, pp. 1--6. 
\bibitem{b6} D. Liu, et al, ``Data-Flow Graph Mapping Optimization for CGRA with Deep Reinforcement Learning,'' in \textit{IEEE Transactions on Computer-Aided Design of Integrated Circuits and Systems}, 2018, pp. 2271--2283. 
\bibitem{b7} Y. Miyasaka, A. Mittal and M. Fujita, ``Synthesis of Algorithm Considering Communication Structure of Distributed/Parallel Computing,'' in \textit{Proceedings of International Symposium on Quality Electronic Design}, 2019, pp. 45--51. 
\bibitem{b8} M. Fujita, Y. Kimura, and Q. Wang, ``Template Based Synthesis
for High Performance Computing,'' in \textit{IFIP/IEEE International Conference on Very
Large Scale Integration (VLSI-SoC)}, 2017, pp. 1-–6. 
\bibitem{b9} https://www.maxeler.com/products/mpc-cseries/ 
\bibitem{b10} A. Goda, et al, ``Synthesis and Generalization of Parallel Algorithms Considering Communication Constraints,'' in \textit{Proceedings of International Symposium on Quality Electronic Design}, 2020, pp. 123--128. 
\bibitem{b11} D. Bahdanau, K. Cho, and Y. Bengio, ``Neural Machine Translation by Jointly Learning to Align and Translate,'' in \textit{International Conference on Learning Representations(ICLR)}, 2015. 
\bibitem{b12} https://pytorch.org/docs/stable/generated/torch.nn.Transformer.html
\bibitem{b13} Berkeley Logic Synthesis and Verification Group, ABC: A System for Sequential Synthesis and Verification, Release 80410. http://www.eecs.berkeley.edu/\~{}alanmi/abc/
\end{thebibliography}
\end{document}